\documentclass[runningheads]{llncs}
\usepackage{graphicx}
\usepackage{multirow}
\usepackage{tabularray}
\usepackage{amsfonts}
\usepackage[breaklinks,colorlinks,citecolor=blue,linkcolor=blue,bookmarks=false]{hyperref}

\begin{document}
\title{MoRA: LoRA Guided Multi-Modal Disease Diagnosis with Missing Modality}
%
%
\author{
Zhiyi Shi\inst{1} 
\and
Junsik Kim\inst{2} 
\and
Wanhua Li\inst{2} 
\and
Yicong Li\inst{2} 
\and
Hanspeter Pfister\inst{2} 
}
\authorrunning{Z. Shi et al.}
\institute{
    Carnegie Mellon University, Pittsburgh PA 15213, USA
    \and
    Harvard University, Cambridge MA 02138, USA\\
    \email{zhiyis@andrew.cmu.edu}
}
\maketitle              
\begin{abstract}
Multi-modal pre-trained models efficiently extract and fuse features from different modalities with low memory requirements for fine-tuning. Despite this efficiency, their application in disease diagnosis is under-explored. A significant challenge is the frequent occurrence of missing modalities, which impairs performance. Additionally, fine-tuning the entire pre-trained model demands substantial computational resources. To address these issues, we introduce Modality-aware Low-Rank Adaptation (MoRA), a computationally efficient method. MoRA projects each input to a low intrinsic dimension but uses different modality-aware up-projections for modality-specific adaptation in cases of missing modalities. Practically, MoRA integrates into the first block of the model, significantly improving performance when a modality is missing. It requires minimal computational resources, with less than 1.6\% of the trainable parameters needed compared to training the entire model. Experimental results show that MoRA outperforms existing techniques in disease diagnosis, demonstrating superior performance, robustness, and training efficiency. The code link is: \href{https://github.com/zhiyiscs/MoRA}{https://github.com/zhiyiscs/MoRA}.

\end{abstract}

\section{Introduction}

Multi-modal pre-trained models achieve immense success in general computer vision tasks including classification and regression \cite{VILT,model1,robust}. Pre-training on extensive and diverse datasets enables multi-modal pre-trained models to understand complex patterns and relationships between different modalities (e.g., images, text, audio, and video). Moreover, the pre-existing knowledge reduces the need for large amounts of task-specific data when adopting these models to a downstream task. 

In recent years, researchers have introduced pre-trained models to medical domains by training multi-modal models on large medical datasets~\cite{medical1,m3ae,HGMF}. However, there are two major challenges in applying these models to disease diagnosis in real clinical settings. Firstly, modality-incomplete situations are quite common in practical disease diagnosis (e.g., patients' chest X-ray images are complete but some of the corresponding annotations are missing). Nevertheless, experiments demonstrate that the performance of multi-modal pre-training models decreases sharply in missing modality situations~\cite{robust}. Secondly, since most of the pre-trained models are based on huge transformers, fine-tuning the entire pre-trained model is still extremely expensive.

Most of the related works~\cite{m3care,SMIL,HGMF} focus on editing the model structure. Nevertheless, these methods can not be applied directly to fine-tuning a pre-trained model. Many works~\cite{m3care,imputation} also adopt imputation in their model, which means imputing a pseudo input for the missing modality based on other complete modalities. However, when the number of modalities is relatively small (e.g. two or three modalities), imputation is extremely non-robust and may even result in worsening outcomes. For fine-tuning multi-modal pre-trained models, Lee \emph{et al.}~\cite{missing_prompt} first introduced the concept of multi-modal prompting, which employs Missing-Aware Prompts (MAPs) to improve the performance when missing modalities are in both the training set and test set. However, MAPs lack robustness in scenarios with different missing modality settings between training and testing. On top of MAPs, Jiang \emph{et al.}~\cite{jiang} proposed Modality-Specific Prompts (MSPs), which are more robust than MAPs towards different missing settings. Nevertheless, MSPs still need to be plugged into several layers to reach the best performance. 

Inspired by Low-rank Adaptation (LoRA)~\cite{LORA}, we propose Modality-aware Low-rank Adaptation (MoRA) to improve the performance and robustness towards missing modalities. Specifically, MoRA operates by projecting each input into a low intrinsic dimension while employing distinct modality-aware up-projections to obtain modality-aware adaptations. These adaptations can discern the unique characteristics of each modality, thereby enhancing the model's robustness and performance in cases where certain modalities are missing. Compared with existing fine-tuning methods, a key advantage of MoRA is its implementation efficiency. It only needs to be integrated into the initial block of the model to lead to significant enhancements in handling missing modalities. During the fine-tuning, all parameters that need to be trained are only MoRA and the classifier. In this way, our method restricts the training parameters to 1.6\% of the total model parameter volume, which allows the model to achieve better performance when fine-tuning on relatively small datasets (several thousand samples). Our experimental results demonstrate that MoRA outperforms existing methods, achieving not only superior accuracy and robustness but also improved training efficiency.
Our main contributions are:
\begin{itemize}
  \item We introduce multi-modal pre-trained models to disease diagnosis and propose MoRA to improve performance and robustness when the data is modality-incomplete in the training and testing sets.
  \item Our method achieves state-of-the-art performance and robustness compared to other fine-tuning methods with missing modalities.
  \item We conduct comprehensive experiments with modalities of different missing rates to demonstrate the superior performance and robustness of our method with different modality-missing ratios.
\end{itemize}

\section{Method}

\subsection{Problem Definition}
\begin{figure}[t]
\centering
\includegraphics[width=\linewidth,height=0.522\linewidth]{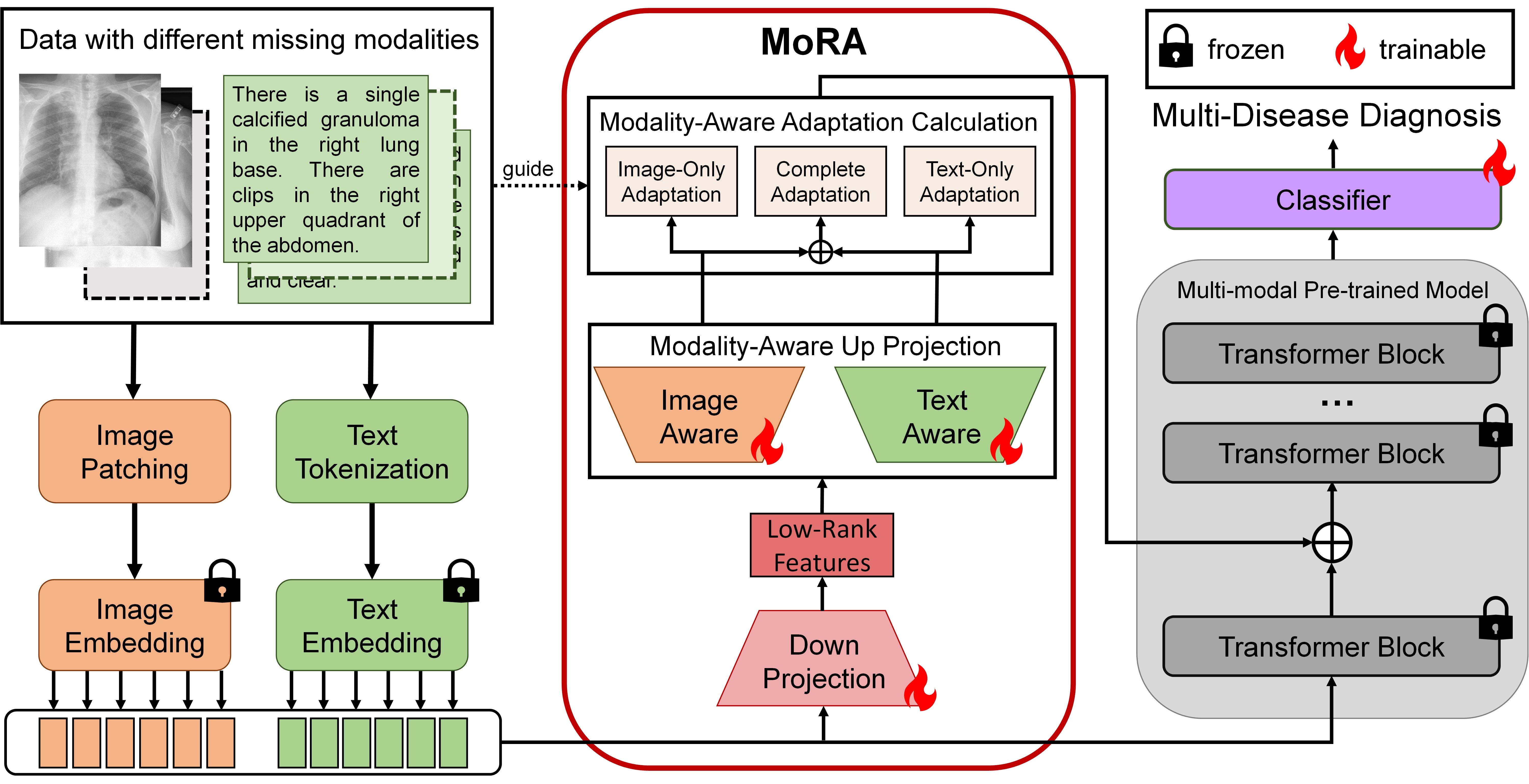}
\caption{The structure of MoRA. Images and texts with different missing modalities are separately embedded into input tokens. MoRA projects these input tokens to a low-rank dimension space and utilizes modality-aware up-projections to obtain modality-aware adaptation. Then, MoRA selects modality-aware adaptation according to the missing case. This adaptation is plugged into the first block of the multi-modal pre-train model (consisting of transformer blocks in our experiments) to extract the features. We feed the output class token to the classifier for multi-disease diagnosis. Trainable parameters are signed by flames while frozen ones are signed by lockers.}
\label{fig1}
\end{figure}

To make it simpler, we take disease diagnosis with two modalities to explain our method. \( m_1 \) and \( m_2 \) (for instance, image and text). We represent this dataset as \( D = \{D^c, D^{m1}, D^{m2}\} \). Here, \( D^c = \{ (x^{m1}_i, x^{m2}_i, y_i) \} \) signifies the subset where both modalities are present, known as the modality-complete subset. Conversely, \( D^{m1} = \{ (x^{m1}_j, y_j) \} \) and \( D^{m2} = \{ (x^{m2}_k, y_k) \} \) represent the modality-incomplete subsets, such as image-only or text-only patients, whose one modality is missing.  As depicted in Figure \ref{fig1}, the dataset comprises complete patients (denoted as \(D^c\)), text-only patients (\(D^{m1}\)), and image-only patients (\(D^{m2}\)). 

To preserve the format of multi-modal inputs for multi-modal pre-trained models, we simply assign dummy inputs \(\tilde{x}^{m1}\), \(\tilde{x}^{m2}\) (e.g., empty string/pixel for texts/images) to the patients with missing modalities and obtain \(\tilde{D}^{m1} = \{x^{m1}_j, \tilde{x}^{m2}_j, y_j\}\), \(\tilde{D}^{m2} = \{\tilde{x}^{m1}_k, x^{m2}_k, y_k\}\). Therefore, the whole patient dataset can be reformed as \(\tilde{D} = \{D^c, \tilde{D}^{m1}, \tilde{D}^{m2}\}\). 

\subsection{Modality-Aware Low-Rank Adaptation}

Low-rank adaptation is widely applied in fine-tuning for large language models. Its main mechanism is to freeze the pre-trained model weights and inject trainable rank decomposition matrices into the pre-trained model. LoRA hypothesizes that the updates to the weights have a low "intrinsic rank" during adaptation. For a pre-trained weight matrix \( W_0 \in \mathbb{R}^{d \times k} \), LoRA constrains its update by representing the latter with a low-rank decomposition \( W_0 + \Delta W = W_0 + BA \), where \( B \in \mathbb{R}^{d \times r} \), \( A \in \mathbb{R}^{r \times k} \), and the rank \( r < \min(d, k) \). During training, \( W_0 \) is frozen and does not receive gradient updates, while \( A \) and \( B \) contain trainable parameters. Note both \( W_0 \) and \( \Delta W = BA \) are multiplied with the same input, and their respective output vectors are summed coordinate-wise. For \( h = W_0x \), the modified forward pass yields:

\begin{equation}
h = W_0x + \Delta W x = W_0x + BAx = W_0x + Adap(x)
\label{eq:lora}
\end{equation}

where \( x \in \mathbb{R}^{1 \times d} \) and \( h \in \mathbb{R}^{1 \times d} \) represent the input and output features, respectively. We consider \( BAx \) as an adaptation for input x ( denoted as \(Adap(x)\)). In practice, rank r is always set to a small number (e.g., 4) and thus LoRA can be trained within limited computation resources. 

On top of LoRA, we propose modality-aware LoRA, which distinguishes from LoRA by introducing modality-aware adaptation. We use a single down-projection \( A \) to project all input \(x \) to the low-rank dimension and get low-rank features \( Ax \). For each modality, we assign a specific modality-aware up-projection (denoted as \( B^{m_1} \in \mathbb{R}^{d \times r}\) and \( B^{m_2} \in \mathbb{R}^{d \times r}\)). After obtaining \(B^{m_1}Ax\) and \(B^{m_2}Ax\), MoRA calculates the adaptation according to the missing case. Specifically, if a patient has data from a certain modality \(m_i\), MoRA then adds the corresponding \(B^{m_i}Ax\) to the adaptation, and vice versa. Thus for the subset \(D^c, \tilde{D}^{m1}, \tilde{D}^{m2}\), their corresponding modality-aware adaptations are:

\begin{equation}
Adap(X^c_i) = (B^{m_1}A+ B^{m_2}A)X_i^c
\label{eq:adap1}
\end{equation}
\begin{equation}
Adap(\tilde{X}_j^{m1}) = B^{m_1}A \tilde{X}_j^{m1}
\label{eq:adap2}
\end{equation}
\begin{equation}
Adap(\tilde{X}_k^{m2}) = B^{m_2}A \tilde{X}_k^{m2}
\label{eq:adap3}
\end{equation}

where \( X^c_i = [x^{m1}_i, x^{m2}_i] \in D^c \), \( \tilde{X}_j^{m1} = [x^{m1}_j, \tilde{x}^{m2}_j] \in \tilde{D}^{m1} \), and \( \tilde{X}_k^{m2} =  [\tilde{x}^{m1}_k, x^{m2}_k]\in \tilde{D}^{m2} \). The selected adaptation will be plugged into the block (which is the first block in our experiment) of multi-modal pre-trained models to improve the robustness towards missing modalities. In the initialization stage, we use a random Gaussian initialization for  \( A \) and zero for \( B^{m_1} \) and \( B^{m_2} \), so adaptation is zero at the beginning of training.

\subsection{Overall Framework}
Following the implementations of~\cite{robust,missing_prompt,jiang}, we utilize the multi-modal pre-trained transformer ViLT~\cite{VILT} as our backbone model, which is designed to deal with two modalities: images and texts. The structure of our method is demonstrated in Fig. \ref{fig1}. Trainable parameters are signed by flames while frozen ones are signed by lockers. 

Patients have images and texts with different missing modalities. For the missing modality, we use a dummy input (which is an empty string for the missing text and a zero matrix for the missing image) to maintain the total number of input tokens for the pre-trained model. We utilize the fixed pre-trained embedding process to transfer data into input tokens. We empirically plug MoRA into the first block (which is a transformer block in ViLT) of the pre-trained model. 

\begin{table}[t]
\centering
    \caption{Splits and types of diseases of our datasets.}
\centering
\resizebox{0.8\linewidth}{!}{
\begin{tblr}{
  cells = {c},
  cell{1}{1} = {r=2}{},
  cell{1}{2} = {c=3}{},
  cell{1}{5} = {r=2}{},
  vline{2-3} = {1-2}{},
  vline{3-5} = {1-2}{},
  vline{2-5} = {3-4}{},
  hline{1,3-5} = {-}{},
  hline{2} = {2-4}{},
}
Datasets & Number of Samples &            &         & Types of Diseases \\
         & Training          & Validation & Testing &                   \\
CXR \cite{CXR}      & 3030              & 385        & 379     & 20                \\
ODIR \cite{ODIR}     & 2781              & 382        & 337     & 7                 
\end{tblr}
}
\label{tab:data}
\end{table}

\section{Experiments}

\subsection{Datasets}
\textbf{Chest X-rays (CXR) Dataset} \cite{CXR} is collected from the open data source of Indiana University. In this dataset, 3794 patients have chest X-ray images, corresponding annotations, and multiple diseases diagnosed by experts. There are a total of 120 different diseases, and we have chosen the top 20 that appear the most as the ones for diagnosis. Note that this dataset contains two CXR image projections: frontal and lateral. In this paper, we only focus exclusively on the frontal projections.

\noindent\textbf{Ocular Disease Intelligent Recognition (ODIR) Dataset} \cite{ODIR} is derived from an ophthalmic database intended to mirror a real-life patient set collected from hospitals. It comprises data from 3,500 patients, specifically curated to aid in the diagnosis of ocular diseases. This dataset encompasses various modalities, including demographic information, clinical text annotations for both eyes and fundus images for each eye.

Splits and types of predicted diseases of our datasets are shown in Table \ref{tab:data}.

\subsection{Implementation Details}
Our code is mainly based on PyTorch and we use PyTorch Lightning for the training and testing inference wrapper. All experiments are conducted on one NVIDIA RTX A4000 GPU. Considering that our model predicts multiple diseases simultaneously, we set a separate binary cross-entropy loss for each disease. 

For MoRA and all the baseline methods, we used the same setting to compare their performance. We freeze all the parameters of ViLT and adopt the same trainable classifier (consisting of two linear layers). We trained the models using AdamW optimization with a batch size of 4 and weight decay of 2e-2. We set the maximum learning rate to 5e-3 and the learning rate was warmed up for 2\% of the total training steps and then decreased linearly to zero. we used the same train, validation, and test splits for every model and trained each model for 40 epochs. If there is no improvement in the results within 5 epochs, the training will be terminated early. We used F1-Macro scores to evaluate the performance of multi-disease prediction.

\subsection{Comparisons with the previous method}

\begin{table}[t]
\centering
    \caption{F1-Macro scores of disease diagnosis on CXR and ODIR test sets with different modality-missing rates. The best value is in \textbf{bold}.}
\resizebox{0.95\linewidth}{!}{
\setlength{\tabcolsep}{4pt}
\renewcommand{\arraystretch}{1.2}
\begin{tabular}{c|cc|cc|cccc} 
\hline
\multirow{2}{*}{Datasets}    & \multicolumn{2}{c|}{Training}                  & \multicolumn{2}{c|}{Test} & \multirow{2}{*}{ViLT \cite{VILT}} & \multirow{2}{*}{MAPs \cite{missing_prompt}} & \multirow{2}{*}{MSPs \cite{jiang}} & \multirow{2}{*}{\begin{tabular}[c]{@{}c@{}}MoRA \\(Ours)\end{tabular}}  \\
                             & Image                  4& text            & Image & text        &                       &                       &                       &                                                                         \\ 
\hline
\multirow{9}{*}{CXR \cite{CXR}} & \multirow{3}{*}{100\%} & \multirow{3}{*}{30\%} & 100\% & 30\%              & 20.47                 & 25.98                 & 26.19                 & \textbf{27.48}                                                                   \\
                             &                        &                       & 30\%  & 100\%             & 25.54                 & 27.78                 & 37.22                 & \textbf{38.19}                                                                   \\
                             &                        &                       & 65\%  & 65\%              & 22.11                 & 22.45                 & 28.69                 & \textbf{30.88}                                                                   \\ 
\cline{2-9}
                             & \multirow{3}{*}{30\%} & \multirow{3}{*}{100\%} & 100\% & 30\%              & 19.78                 & 30.90                & 33.49                & \textbf{35.13}                                                                   \\
                             &                        &                       & 30\%  & 100\%             & 50.32                 & 71.23                 & 72.98                 & \textbf{75.37}                                                                    \\
                             &                        &                       & 65\%  & 65\%              & 23.34                 & 51.62                & 52.79                 &\textbf{54.69}                                                                   \\ 
\cline{2-9}
                             & \multirow{3}{*}{65\%} & \multirow{3}{*}{65\%} & 100\% & 30\%              & 33.89                & 36.87                 & 37.09                 & \textbf{37.78}                                                                 \\
                             &                        &                       & 30\%  & 100\%             & 57.51                 & 67.57                 & 67.47                 & \textbf{68.99}                                                                   \\
                             &                        &                       & 65\%  & 65\%              & 36.51                & 52.32                 & 52.57                 & \textbf{54.63}                                                                   \\ 
\hline
\multirow{9}{*}{ODIR \cite{ODIR}}        & \multirow{3}{*}{100\%} & \multirow{3}{*}{30\%} & 100\% & 30\%              & 50.18                 & 58.43                 & 58.96               & \textbf{60.94}                                                                   \\
                             &                        &                       & 30\%  & 100\%             & 81.34                 & 46.38                 & 90.66                 & \textbf{92.56}                                                                   \\
                             &                        &                       & 65\%  & 65\%              & 67.26                 & 46.74                 & \textbf{78.71}                 & 76.89                                                                 \\ 
\cline{2-9}
                             & \multirow{3}{*}{30\%} & \multirow{3}{*}{100\%} & 100\% & 30\%              & 46.21                 & 50.56                 & 45.98                 & \textbf{58.69}                                                                   \\
                             &                        &                       & 30\%  & 100\%             & 95.53                 & 98.95                 & 99.22                & \textbf{99.77}                                                                   \\
                             &                        &                       & 65\%  & 65\%              & 75.61                 & 78.54                 & 77.21                & \textbf{79.61}                                                                   \\ 
\cline{2-9}
                             & \multirow{3}{*}{65\%} & \multirow{3}{*}{65\%} & 100\% & 30\%              & 57.28                 & 57.84                 & 58.98                 &     \textbf{59.00}                                                            \\
                             &                        &                       & 30\%  & 100\%             & 96.82                 & 98.40                 & 99.33                 & \textbf{99.54}                                                                     \\
                             &                        &                       & 65\%  & 65\%              & 77.11                 & 78.60                 & 78.94                 & \textbf{80.73}                                                                   \\
\hline
\end{tabular}
}
 \label{tab:comparison}
\end{table}

In this part, we conduct experiments under different missing settings in the training set and testing set to compare MoRA with three previous methods. The experimental results of F1-Macro are shown in Table \ref{tab:comparison}. It can be observed that MoRA achieves the best results in most of the missing scenarios even when missing rates are not the same in the training set and the test set. It is worth mentioning that, according to the settings of the original MSPs and MAPs papers, we have inserted them into the first to sixth blocks, whereas we only inserted MoRA into the first block. Results demonstrate that MoRA can achieve better performance by plugging into only the first block. It can also be seen from the table that the model's robustness to text is significantly weaker than to image. This is reasonable in practical multi-modal learning: the importance of one modality is greater than that of others. Thus, how to improve the robustness of this important modality is crucial. It can be observed that MoRA performs much better when text is seriously missing.

\begin{table}[t]
\centering
\caption{GPU requirements and total time for 1000 training steps.}
\resizebox{0.9\linewidth}{!}{
\setlength{\tabcolsep}{4pt}
\renewcommand{\arraystretch}{1.2}
\begin{tabular}{c|cc|cc} 
\hline
             & \multicolumn{2}{c|}{  CXR \cite{CXR}}                             & \multicolumn{2}{c}{  ODIR \cite{ODIR}}                              \\
             & GPU Memory                                  & Training Time                              & GPU Memory                                  & Training Time                               \\ 
\hline
MAPs \cite{missing_prompt}         & 14.4 GB                                     &  {1.71 h} &   {13.0 GB} &   {1.82 h}  \\
MSPs \cite{jiang}         &   {12.4 GB} &   {1.75 h} &   {12.1 GB} &   {1.85 h}  \\
MoRA (Ours)~ &   {12.2 GB} &   {1.58 h} &   {11.6 GB} &   {1.59 h}  \\
\hline
\end{tabular}
}
\label{lab:GPUs}
\end{table}

We also compared the GPU memory requirements and training times of different methods during training. As shown in Table \ref{lab:GPUs}, at 1000 training steps, MoRA requires relatively smaller GPU memory and shorter training time. This is because MoRA only needs to be inserted into the first layer of the pre-trained model, resulting in fewer trainable parameters.

\subsection{Ablation Study}

\begin{figure}[t]
\centering
\includegraphics[width=0.7\linewidth]{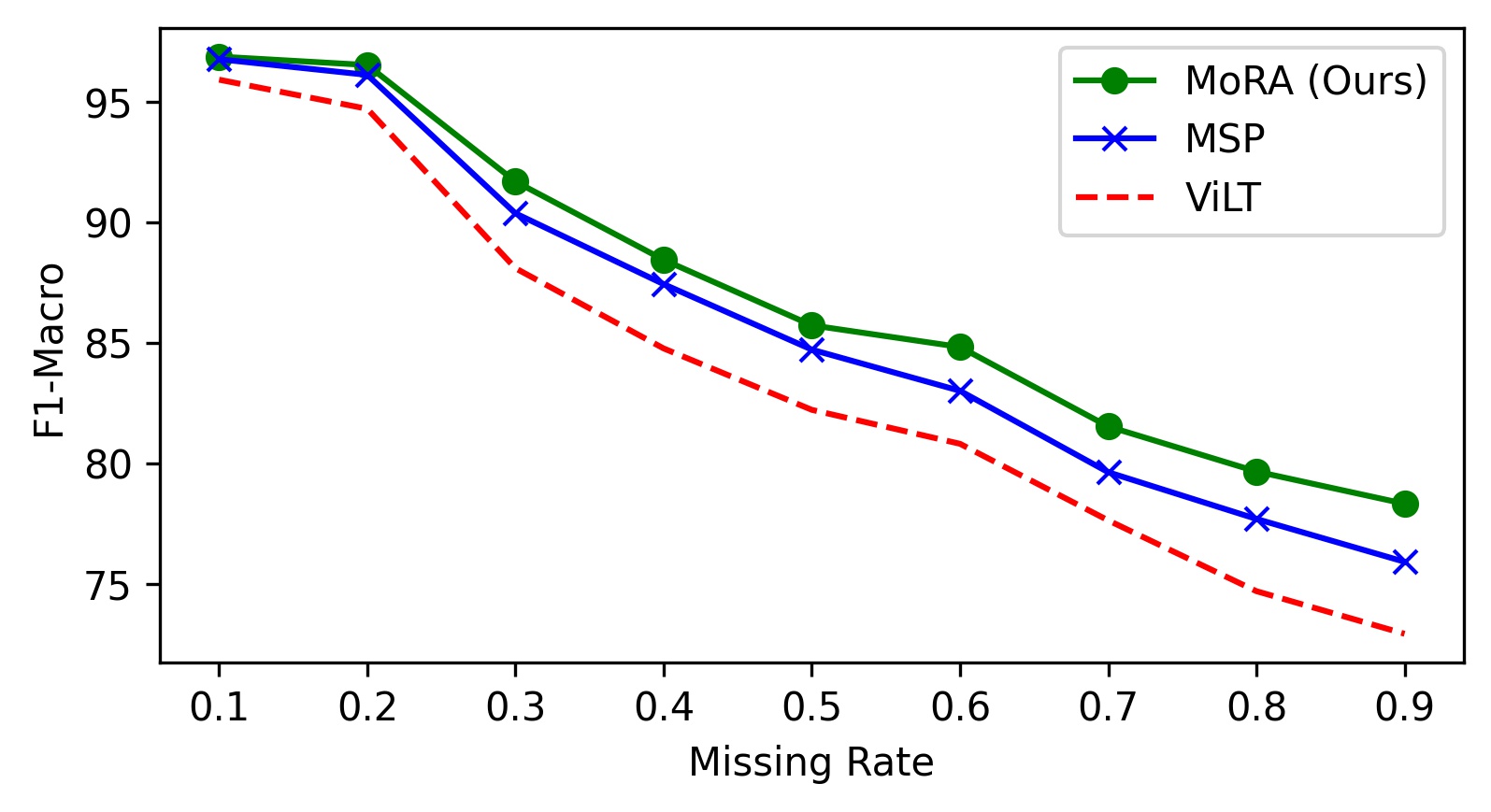}
\caption{F1-Macro scores on ODIR with different missing rates.}
\label{fig2}
\end{figure}

\noindent\textbf{Robustness to different missing rates.} We conduct further experiments to analyze the robustness of our proposed
method against different missing-modality rates. To make it clear, we maintained the missing rates for each modality are the same and we consider the total missing rate as \(\eta\). We trained MoRA on ODIR with  \(\eta=70\%\), which means 65\% image-modality and 65\% text-modality samples. We tested it with different missing rates and demonstrated results in Figure \ref{fig2}. When the missing rate is small, our method and baseline results are not significantly different. As the missing rate continues to increase, our model exhibits greater robustness. This indicates that our model can more effectively cope with extreme modal missing situations.

\begin{table}[t]
\centering
\caption{F1-Macro scores with different plugged blocks.}
\resizebox{0.9\linewidth}{!}{
\setlength{\tabcolsep}{10pt}
\renewcommand{\arraystretch}{1.2}
\begin{tabular}{c|ccccc} 
\hline
Plugged Blocks & {[}0] & {[}0,1] & {[}0,1,2] & {[}0,1,2,3,4,5] & {[}5]  \\ 
\hline
MoRA           & 80.73 & 80      & 80.89     & 80.68           & 78.48  \\ 
\hline
MAP            & 74.58 & 75.01   & 75.56     & 78.6            & 76.86  \\
\hline
\end{tabular}
}
\label{lab:block}
\end{table}

\noindent\textbf{The effect of plugged blocks.} According to \cite{missing_prompt}. MAPs are quite sensitive to the plugged blocks.  We also conduct experiments to analyze the effect of locations on MoRA. We trained MoRA on ODIR with65\% image-modality and 65\% text-modality samples and fixed the rank r. We try to plug MoRA into different blocks to check the performance. According to Table \ref{lab:block}, experiments show that the performance of plugging MoRA into several blocks is close to plugging into the first block. It can be observed that compared to MAPs, MoRA is not very sensitive to the number of inserted blocks. However, in the experiments, we found that inserting into the first block is crucial for the effectiveness of MoRA. This may be because the first layer can directly obtain information about the input token, which can help MoRA confirm the status of missing modularity and facilitate subsequent fine-tuning. So in practical use, MoRA is most suitable to be inserted into the first block to help fine-tuning, which can use as few training parameters as possible based on obtaining good results. This is also the advantage of MoRA compared to MAPs.

\begin{table}[t]
\centering
\caption{F1-Macro scores on ODIR with different rank r.}
\resizebox{0.9\linewidth}{!}{
\setlength{\tabcolsep}{10pt}
\renewcommand{\arraystretch}{1.2}
\begin{tabular}{c|c|c|c|c|c|c} 
\hline
Rank r   & 1   & 2   & 4   & 16  & 32 & 384  \\ 
\hline
F1-Macro & 77.52 & 78.98 & 80.73 & 80.59 & 80.32 & 70.23 \\
\hline
\end{tabular}
}
\label{lab:rank}
\end{table}

\noindent\textbf{The effect of rank r.} We examine the impact of rank r on performance. We trained MoRA on ODIR with 65\% image-modality and 65\% text-modality samples but set different rank r. We demonstrate results in Table \ref{lab:rank}. As the table shows, the performance is getting better when the rank is increasing. However, results indicate that optimal performance is achieved when the rank is set to 4. We also tested an extreme scenario where rank r equals the dimension of the input token. In this case, the results are very poor, even worse than cases without MoRA. This indicates that MoRA can only play a role when the rank is very small, which is consistent with the derivation of LoRA. Overall, MoRA is not very sensitive to the choice of r.

\section{Conclusion}

In this paper, we introduce multi-modal pre-trained models for disease diagnosis. To tackle the challenges, We propose MoRA for fine-tuning multi-modal pre-trained models with missing modalities. MoRA projects each input to the same low intrinsic dimension but utilizes different modality-aware up-projections to obtain modality-aware adaptation for certain modality-missing cases. We conduct experiments on two disease diagnosis tasks with different modality-missing rates, which demonstrate the advantages of our method. MoRA can not only improve the robustness and performance but also save computational resources. 
In future work, we will extend our method to larger pre-trained models and explore the feasibility of introducing large multi-modal pre-trained models into disease diagnosis. 

\begin{credits}
\subsubsection{\ackname}
We thank all affiliates of the Harvard Visual Computing Group for their valuable feedback. This work was supported by NIH grant R01HD104969 and NIH grant 1U01NS132158.

\subsubsection{\discintname}
The authors have no competing interests to declare that are relevant to the content of this article.
\end{credits}

%
%
%
%

\end{document}